\documentclass[nai]{iosart2x}

\pubyear{0000}
\volume{0}
\firstpage{1}
\lastpage{1}
\usepackage{todonotes}
\usepackage{multirow}
\usepackage{mathtools}
\usepackage{graphicx}
\usepackage[english]{babel}
\usepackage{amsthm}
\usepackage{booktabs}
\usepackage{multirow}
\usepackage{amssymb}
\usepackage{pifont}
\usepackage{xspace}
\usepackage{comment}
\newtheorem{definition}{Definition}
\newcommand{\cmark}{\ding{51}\xspace}%
\newcommand{\xmark}{\ding{55}\xspace}%

\newcounter{todocounter}
\newcommand{\ma}[1]{\stepcounter{todocounter}
  {\color{blue} Mehwish: \thetodocounter: #1}}

\usepackage{comment}

\begin{document}

\begin{frontmatter}

\title{Towards Semantically Enriched Embeddings for Knowledge Graph Completion}

\runtitle{Towards Semantically Enriched Embeddings for Knowledge Graph Completion}


\begin{aug}
\author[A]{\inits{M.}\fnms{Mehwish} \snm{Alam}\ead[label=e1]{mehwish.alam@telecom-paris.fr}%
\thanks{Corresponding author. \printead{e1}.}}
\author[B]{\inits{F.}\fnms{Frank} \snm{van Harmelen}\ead[label=e2]{frank.van.harmelen@vu.nl}}
\author[C]{\inits{M.}\fnms{Maribel} \snm{Acosta}\ead[label=e3]{maribel.acosta@tum.de}}

\address[A]{\orgname{Telecom Paris, Institut Polytechnique de Paris}, \cny{France}\printead[presep={\\}]{e1}}
\address[B]{\orgname{Vrije Universiteit Amsterdam}, \cny{the Netherlands}\printead[presep={\\}]{e2}}
\address[C]{School of Computation, Information and Technology, \orgname{Technical University of Munich}, \cny{Germany}\printead[presep={\\}]{e3}}
\end{aug}






\begin{abstract}
Embedding based Knowledge Graph (KG) completion has gained much attention over the past few years. Most of the current algorithms consider a KG as a multidirectional labeled graph and lack the ability to capture the semantics underlying the schematic information. 
This position paper revises the state of the art and discusses several variations of the existing algorithms for KG completion, which are discussed progressively based on the level of expressivity of the semantics utilized. The paper begins with analysing various KG completion algorithms considering only factual information such as transductive and inductive link prediction and entity type prediction algorithms. It then revises the algorithms utilizing Large Language Models as background knowledge.
Afterwards, it discusses the algorithms progressively utilizing semantic information such as class hierarchy information within the KGs and semantics represented in different description logic axioms.
The paper concludes with a critical reflection on the current state of work in the community, where we argue that the aspects of semantics, rigorous evaluation protocols, and bias against external sources have not been sufficiently addressed in the literature, which hampers a more thorough understanding of advantages and limitations of existing approaches. Lastly, we provide recommendations for future directions.
\end{abstract}


\end{frontmatter}




\vspace{-1.5cm}

\section{Introduction}\label{sec:introduction}
Knowledge Graphs (KGs) have recently gained attention due to their applicability to diverse fields of research, ranging from knowledge management, representation, and reasoning to learning representations over KGs. 
KGs represent knowledge in the form of relations between entities structured as (head, relation, tail) triples, referred to as facts, along with schematic information in the form of ontologies. KGs have been used for various downstream tasks such as web search, recommender systems, and question answering. 

KGs, however, suffer from incompleteness because of manual or automated generation. Manual generation leads to limited knowledge represented by the curator and contains curator bias~\cite{ntoutsi2020bias}, while automated generation may lead to erroneous or missing information. 
KG completion in particular includes the tasks of (i) triple classification, i.e., deciding if the triple is true or not, (ii) Link Prediction (LP) to complete the head, tail, or relation in a triple, and (iii) entity classification, also known as entity type prediction. 
To perform KG completion, various rule-based as well as embeddings based models have been proposed. 
These algorithms are computationally expensive and are transductive: they only perform predictions based on triple information involving known entities and relations. This is not readily usable when the inference has to be performed on unseen entities and relations. 
To this end, inductive KG completion allows for predicting triples that involve unseen entities and relations. 
These transductive and inductive LP algorithms are mostly based on factual information contained in KGs. Various studies leveraging language models as an external source of knowledge have been proposed for KG completion. These algorithms lag behind in terms of performance w.r.t. KG embedding based methods because of ranking-based metrics such as \textit{Hits@k} since existing KG embedding-based algorithms operate under the Closed World Assumption. 
This led to the need for human evaluation since Large Language Models (LLMs) contain more general knowledge and may generate correct answers that are different from what is expected by the ground truth with the highest score. Apart from the factual information (i.e., Assertion Box, ABox), another source of information is the ontological information (i.e., Terminology Box, TBox) contained in the KG. The current methods almost completely ignore this TBox information.  
To rectify this situation various attempts have been made to include type hierarchies and ontological information with different expressivity levels such as $\mathcal{EL}^{++}$, $\mathcal{ALC}$, etc. In some cases, additional representational capabilities are utilized to capture this information such as box embeddings.

\paragraph{Related Work.} 
Several studies have surveyed the state of the art (SoTA) in KG completion. 
The work by Paulheim~\cite{Paulheim17} provides a survey of the articles related to KG refinement including various classical and rule-based approaches for KG completion, yet KG embeddings are not discussed. 
Other surveys specifically focus on KG embedding-based methods for KG completion. 
Wang et al.~\cite{WangMWG17} organises the algorithms for embedding-based KG completion according to their scoring functions such as translational models, semantic matching models, etc. However, this survey does not discuss the methods proposed for KG completion using multimodal information related to an entity or relation such as images, text, and numeric literals. 
These aspects are targeted in the survey by Gesese et al.~\cite{DBLP:journals/semweb/GeseseBAS21}, which categorizes these methods based on their scoring function (inspired by the work by Wang et al.~\cite{WangMWG17}) as well as based on multiple modalities. The survey shows theoretical and experimental comparisons of the existing approaches. 
Still, none of the aforementioned works cover the aspects of ontological statements and semantics in KG embeddings. 
Recent works~\cite{pan_et_al:TGDK.1.1.2,pan2024unifying}  discuss the interplay of KGs and LLMs. 
The study by S. Pan et al.~\cite{pan2024unifying} describes methods that integrate LLMs for KG embeddings and KG completion, yet the role of schematic information is not considered in the anlaysis. 
Complementary, J. Z. Pan et al.~\cite{pan_et_al:TGDK.1.1.2} describe how LLMs can be used to complete triples and perform different ontology engineering tasks (e.g., completion or refinement) in KGs.  
In contrast to these works, our work focuses on KG embeddings and how LLMs can be used as external sources to improve the performance of KG completion tasks.
In summary, as compared to these existing studies, the current article targets the semantic aspects of knowledge graph embeddings by summarizing and discussing the approaches that have been proposed so far for leveraging semantics provided in the KG.

\paragraph{Contributions.} 
This position paper provides an overview of the evolution of methodologies proposed for KG completion, starting from the embedding-based algorithms and LLM-based approaches to the various categories of algorithms proposed for incorporating schematic information within KG embeddings for performing different kinds of completion tasks. 
It further discusses the limitations of the approaches in each of these categories and concludes with critical reflections and future research directions.

\section{Preliminaries: Semantics in Knowledge Graphs}\label{sec:preliminaries}

Commonly, KGs are defined as a set of head-relation-tail triples $(h,r,t)$, where $h$ and $t$ are nodes in a graph, and $r$ is a directed edge from $h$ to $t$ (e.g., \cite{bordes2013translating,nickel2011three,dettmers2018conve,jia2020improving}). 
In this way, a KG corresponds to a directed, labeled graph, where triples in the KG are labeled edges between labeled nodes, as captured in the following definition. 

\begin{definition}[Knowledge Graph: Triple Set Definition]
\label{def:triple-set}
A knowledge graph is a directed, labeled graph $G=(V,E,l, L_G)$, with $V$ the set of nodes, $E$ the set of edges, $L_G$ the set of labels, and a mapping $l : V \cup E \rightarrow L_G$. 
A triple $t=(h,r,t)$ is a labelled edge, i.e., $(h,r,t)=(l(h), l(r), l(t))$ where $r \in E$ and $h, t \in V$.  
\end{definition}

\begin{table}[]
    \centering
    \caption{Axioms in $\mathcal{EL}^{++}$, $\mathcal{ALC}$, and $\mathcal{SH}$. Notation: $\top$ top, $\bot$ bottom, $a,b$ are instances, $C,D$ are concepts, $r, s$ are relations, $trans$ denotes transitivity.}
    \label{tab:logic}
\begin{tabular}{lllllllllllllll}
\toprule
                    & $\top$     & $\bot$     & $\{a\}$        & $r(a,b)$   & $C(a)$       & $C \sqcap D$ & $C \sqsubseteq D$ & $C \sqcup D$ & $\neg C$   & $\exists$  & $\forall$  & $r \sqsubseteq s$ &   $trans(r)$ & $r_1 \circ \ldots \circ r_n \sqsubseteq s$  \\
\midrule
$\mathcal{EL}^{++}$ & \cmark & \cmark & \cmark & \cmark & \cmark & \cmark   & \cmark        & \xmark   & \xmark & \cmark & \xmark & \xmark        & \xmark &\xmark  \\
$\mathcal{ALC}$     & \cmark & \cmark & \cmark & \cmark & \cmark & \cmark   & \cmark        & \cmark   & \cmark & \cmark & \cmark & \xmark        & \xmark & \xmark  \\
$\mathcal{SH}$      & \cmark & \cmark & \cmark & \cmark & \cmark & \cmark   & \cmark        & \cmark   & \cmark & \cmark & \cmark & \cmark        & \cmark &\cmark  \\
\bottomrule
\end{tabular}
\end{table}

Definition~\ref{def:triple-set} captures the graph structure of KGs, yet aspects concerning the meaning of nodes and edges in a KG are not explicitly defined. 
First, there is no distinction about the kind of nodes that can exist in a KG, i.e., whether nodes correspond to entities or text descriptions.  
This distinction exists in data models like the Resource Description Framework (RDF), where nodes can represent entities, labeled with IRIs or blank nodes, or literals that are used for values such as strings or different types of numbers like integers, float, etc. 
Distinguishing between entities and literals impacts both the structure and connectivity of nodes in the graph as well as the meaning of things in the graph. 
As we will see in Section~\ref{sec:transductive-link-prediction}, specialised KG embedding models are required to handle both kinds of nodes and different types of literals in a KG.    
Second, the semantics (formal meaning) of classes or relations in the KG are not provided in Definition~\ref{def:triple-set}.  
This is typically the role of ontologies in KGs,  which specify the definitions for classes (or concepts) and relations (or roles)  using labels or symbols coming from logic.  
Different logical languages introduce different levels of expressivity (cf. Table~\ref{tab:logic}). 
Expressivity here refers to the allowed complexity of statements that go beyond simple assertions about individuals (e.g., class assertions) and their relations to other individuals or literals, that are comprised in the ABox. 
More expressive statements included in the TBox comprise class and role definitions. 
The constructs and axioms defined in a logic language $\mathcal{L}$ can be transformed into labels and triples to be encoded in a KG. 
Based on this, we provide a definition for KGs that takes into account the semantics of statements.      

\begin{definition}[Knowledge Graph: Semantic Definition]
\label{def:semantic-kg}
Let $\mathcal{L}$ be a logic language that defines the semantics of concepts and roles, and  $G=(V,E,l,L_G)$ a knowledge graph (KG) as of Definition~\ref{def:triple-set}. 
$G$ is an $\mathcal{L}$-KG if the labels $L_G$ contain all the symbols defined in $\mathcal{L}$, and the triples in $G$ correspond to statements that can be expressed in $\mathcal{L}$.  
\end{definition}

Typical logic languages used in KGs are in the family of Description Logics, due to their convenient trade-off between expressivity and scalability~\cite{rudolph2011foundations}.  
For example, the existential language $\mathcal{EL}$ defines the notions of concept intersection and full existential quantification. 
The extension $\mathcal{EL}^{++}$ introduces the additional notions of concept intersection and concept subsumption; the latter is necessary to model class hierarchies in KGs.  
$\mathcal{ALC}$ provides additional expressivity w.r.t. $\mathcal{EL}^{++}$ as it includes concept union, negation, and universal quantification. 
Higher levels of expressivity are captured in KGs modeled with formalisms defined for the semantic web like the Web Ontology Language (OWL). 
OWL is based on the $\mathcal{SH}$ language, which includes more complex definitions for roles, including role subsumption and transitivity.
Another important aspect of KG semantics is the concept of data types captured in the $(\mathcal{D})$ extension. 
$(\mathcal{D})$ allows for modeling the meaning of literals in KGs which can be part of statements in the ABox. 
It is important to note that KGs based on OWL can achieve different levels of expressivity beyond the ones described in this section, 
e.g., OWL-Lite is based on  $\mathcal{SHIF^{(D)}}$, OWL-DL on $\mathcal{SHOIN^{(D)}}$,  and OWL2 on $\mathcal{SROIQ^{(D)}}$, the latter providing definitions for role reflexivity,  irreflexivity, and disjointness, inverse properties, enumerated classes, and qualified cardinality restrictions.

\section{Knowledge Graph Completion using Embeddings}\label{sec:kg-embedding}
The vast majority of embedding based algorithms for KG completion treat KGs as graph structures as presented in Definition~\ref{def:triple-set}.
Most of these algorithms perform transductive link prediction (LP)~\cite{dai2020survey} where the inference is performed on the same graph on which the model is trained.  
On the contrary, inductive LP is performed over the unseen graph, i.e., (parts of) the graph are not seen during training. This section gives an overview of KG embedding algorithms for transductive LP (Section~\ref{sec:transductive-link-prediction}), inductive LP algorithms (Section~\ref{sec:inductive-link-prediction}), and entity type prediction (Section~\ref{sec:entity-type-prediction}) since the previous two categories mostly focus on head, tail, relation, and triple prediction.

\subsection{Transductive Link Prediction}
\label{sec:transductive-link-prediction}

A large variety of KG embeddings have been proposed for the LP task, which differ in their scoring function (which measures the plausibility of a triple in a KG) and their underlying learning methodology, such as translational models, semantic matching models, neural network models, path-based methods, and models based on multimodal KG -- e.g., making use of textual, numeric, and multi-modal literals. 
In the following, a brief overview of the models performing transductive LP is given. A detailed description of these models is provided by Dai et al.~\cite{dai2020survey}.

\textit{Translation based models} include TransE, TransH, etc. In TransE~\cite{bordes2013translating}, the relation in a triple is considered a translation operation between the head and tail entities on a low dimensional space. 
TransH~\cite{wang2014knowledge} extends TransE by projecting the entity vectors to relation-specific hyperplanes which helps in capturing different roles of an entity w.r.t. different relations. 
Both models have obvious limitations, such as the inability to represent symmetric or transitive relations. 
The scoring function of RotatE~\cite{DBLP:conf/iclr/SunDNT19} models the relation as a rotation in a complex plane to preserve the symmetric/anti-symmetric, inverse, and composition relations in a KG. 


\textit{Semantic matching models} are based on a similarity-based scoring function that measures the plausibility of a triple by matching the semantics of the latent representations of entities and relations.
In DistMult~\cite{corr/YangYHGD14a}, each entity is mapped to a $d-$dimensional dense vector, and each relation is mapped to a diagonal matrix. The score of a triple is computed as the matrix multiplication between the entity vectors and the relation matrix. RESCAL~\cite{nickel2011three} models the triple into a three-way tensor. 
The model explains triples via pairwise interaction of latent features. The score of the triple is calculated using the weighted sum of all the pairwise interactions between the latent features of the head and the tail entities. ComplEx~\cite{Trouillon2016ComplEx} extends DistMult by introducing a Hermitian dot product for better handling asymmetric relations. 

\textit{Neural network models} represent an entity using the average of the word embeddings in the entity name. ConvE~\cite{dettmers2018conve} uses 2D convolutional layers to learn the embeddings of the entities and relations in which the head entity and the relation embeddings are reshaped and concatenated which serves as an input to the convolutional layer. The resulting feature map tensor is then vectorized and projected into a k-dimensional space and matched with the tail embeddings using the logistic sigmoid function minimizing the cross-entropy loss. In ConvKB~\cite{dai2018novel}, each triple is represented as a 3-column matrix which is then fed to a convolution layer. Multiple filters are applied to the matrix in the convolutional layer to generate different feature maps. Next, these feature maps are concatenated into a single feature vector representing the input triple. The feature vector is multiplied with a weight vector via a dot product to return a score which is used to predict whether the triple is valid. 
Relational Graph Convolutional Networks (R-GCN)~\cite{schlichtkrull2018modeling} extends Graph Convolutional Networks (GCN) to distinguish different relationships between entities in a KG. In R-GCN, different edge types use different projections, and only edges of the same relation type are associated with the same projection weight. 


\textit{Path-based models} such as PTransE~\cite{DBLP:conf/emnlp/LinLLSRL15} extend TransE by introducing a path-based translation model. GAKE~\cite{GAKE} considers the contextual information in the graph by considering the path information starting from an entity. RDF2Vec~\cite{ristoski2016rdf2vec} uses random walks to consider the graph structure and then applies the word embedding model on the paths to learn the embeddings of the entities and the relations. However, the prediction of head or tail entities with RDF2Vec is non-trivial because it is based on a language modeling approach. 

\textit{Multimodal KG embeddings} make use of different kinds of literals such as numeric, text, or images (see~\cite{DBLP:journals/semweb/GeseseBAS21} for detailed analysis). 
This group of algorithms considers aspects of the $\mathcal{(D)}$ extension of logic languages (cf. Section~\ref{sec:preliminaries}). 
For example, DKRL~\cite{xie2016representation} extends TransE by incorporating the textual entity descriptions encoded using a continuous bag-of-words approach. Jointly (ALSTM)~\cite{xu2017knowledge} extends the DKRL model with a gating strategy and uses attentive LSTM to encode the textual entity descriptions. MADLINK~\cite{madlink} uses SBERT for representing entity descriptions and the structured representation is learned by performing random walks where at each step the relations to be crawled are ranked using ``predicate frequency - inverse triple frequency'' \textit{(pf-itf)}.


\subsection{Inductive Link Prediction}
\label{sec:inductive-link-prediction}

Simply adapting most of the existing transductive LP models for inductive settings requires expensive re-training for learning embeddings for unseen entities leading to their inapplicability to perform predictions with unseen entities. To overcome this, inductive LP approaches were introduced which are discussed in the following. 


\textit{Statistical rule-mining approaches} make use of logical formulas to learn patterns from KGs. 
Systems such as AnyBURL~\cite{MeilickeCRS19,Meilicke2018FineGrainedEO} generalise random walks over the graph into Horn Clause rules which are then used for link-prediction: if the body of the rule matches with a path in the graph, the rule predicts that the triple in the conclusion of the rule should also be in the graph.  
NeuralLP~\cite{yang2017NeuralLP} was proposed which learns first-order logical rules in an end-to-end differentiable model. DRUM~\cite{Sadeghian2019DRUM} is another method that applies a differentiable approach for mining first-order logical rules from KGs and provides an improvement over NeuralLP.

\textit{Embedding based methods} have also been proposed to work in an inductive setting. 
GraphSAGE~\cite{Hamilton2017GraphSage} performs inductive LP by training entity encoders through feed-forward and graph neural networks. However, in GraphSAGE, the set of attributes (e.g., bag-of-words) are fixed before learning entity representations, restricting their application to downstream tasks~\cite{Daza2021InductiveER}.  
One way to learn entity embeddings is to use graph neural networks for aggregating the neighborhood information~\cite{Hamaguchi2017transfer, Wang2019LAN}. However, these methods require unseen entities to be surrounded by known entities and fail to handle entirely new graphs~\cite{Teru2020GraIL} (i.e. they work only in a semi-inductive setting). KEPLER~\cite{Wang2021KEPLERAU} is a unified model for knowledge embedding and pre-trained language representation by encoding textual entity descriptions with a pre-trained language model (LM) as their embeddings, and then jointly optimizing the KG embeddings and LM objectives. However, KEPLER is computationally expensive due to the additional LM objective and requires more training data. Inspired by DKRL, BLP~\cite{daza2021inductive} uses a pre-trained LM for learning representations of entities via an LP objective. QBLP~\cite{Ali2021ImprovingIL} is a model proposed to extend BLP for hyper-relational KGs by exploiting the semantics present in qualifiers. Previously discussed models only consider unseen entities and not unseen relations. RAILD~\cite{GeseseSA22} on the other hand generates a relation-to-relation network for efficiently learning the relation features leading to solving the task of fully-inductive LP for unseen relations. RMPI~\cite{GengCPCJZC23} is another method that performs full-inductive link prediction by using relational message passing whereas the traditional methods perform message passing over the entities. A detailed survey on inductive LP algorithms is provided by Hubert et al.~\cite{corr/abs-2312-04997}.

\subsection{Entity Type Prediction}
\label{sec:entity-type-prediction}


SDType~\cite{paulheim2013type} 
is a statistical heuristic model that exploits links between instances using weighted voting and assumes that certain relations occur only with particular types. It does not perform well if two or more classes share the same sets of properties and also if specific relations are missing for the entities.
Many machine learning including neural network based models have been proposed for type prediction.  Cat2Type~\cite{BiswasSSA21} takes into account the semantics of the textual information in Wikipedia categories using language models such as BERT. In order to consider the structural information of Wikipedia categories, a category-to-category network is generated which is then fed to Node2Vec for obtaining the category embeddings. The embeddings of both structural and textual information are combined to classify entities into their types. The approach by Biswas et al.~\cite{BiswasPPSA22} leverages different word embedding models, trained on triples, together with a classification model to predict the entity types. 
Therefore, contextual information is not captured. Scalable Local Classifier per Node (SLCN) is used by the model in~\cite{melo2016type} for type prediction based on a set of incoming and outgoing relations. However, entities with only a few relations are likely to be misclassified.  FIGMENT~\cite{YaghoobzadehAS18} uses a global model and a context model. The global model predicts entity types based on the entity mentions from the corpus and the entity names. The context model calculates a score for each context of an entity and assigns it to a type. Therefore, FIGMENT requires a large annotated corpus which is a drawback of the method. In APE~\cite{jin-etal-2018-attributed}, a partially labeled attribute entity-to-entity network is constructed containing structural, attribute, and type information for entities, followed by deep neural networks to learn the entity embeddings. MRGCN~\cite{wilcke2020end} is a multi-modal message-passing network that learns end-to-end from the structure of KGs as well as from multimodal node features. In HMGCN~\cite{jin-etal-2019-fine}, the authors propose a GCN-based model to predict the entity types considering the relations, textual entity descriptions, and Wikipedia categories. ConnectE~\cite{zhao2020connecting} and AttET~\cite{zhuo2022neighborhood} models find a correlation between neighborhood entities to predict the missing types. Ridle~\cite{weller2021predicting} learns entity embeddings and latent distribution of the relations using Restricted Boltzmann Machines allowing to capture semantically related entities based on their relations. This model is tailored to KGs where entities from different classes are described with different relations.
CUTE~\cite{XuZLXHW16} performs hierarchical classification for cross-lingual entity typing by exploiting category, property, and property-value pairs. MuLR~\cite{SchutzeY17} learns multi-level representations of entities via character, word, and entity embeddings followed by the hierarchical multi-label classification.

\paragraph{Discussion.}
The algorithms discussed in Section~\ref{sec:kg-embedding}
focus on encoding the factual information and structural relationships among entities in the KG. 
In other words, these approaches treat KGs as a set of triples and consider only statements in the ABox for generating KG embeddings and performing LP.
Despite that some embeddings are computed not only triple-level information but also contextual entity information based on their neighboring relations and nodes (by implementing graph walks), all these approaches do not consider other forms of knowledge, particularly those found in LLMs or in expressive axioms of logical languages that form the backbone of knowledge graph semantics. 
For this reason, despite the great advances in reasoning achieved by KG embeddings, they still lack a deeper semantic understanding of entities, relationships, and classes in KGs and the more complex and logical constraints that underlie them.
The subsequent sections focus on these aspects of LP.  

\section{Large Language Models for Knowledge Graph Completion}\label{sec:llm-kg-completion}

LLMs can further be categorized into encoder-decoder models, encoder-only, and decoder-only models. The encoder-decoder models and encoder-only models, such as BERT~\cite{DevlinCLT19}, RoBERTa~\cite{corr/abs-1907-11692}, etc., are masked language models which are discriminative models and are pretrained for predicting a masked word. These models have achieved SoTA performance on the NLP tasks such as entity recognition, sentiment analysis, etc. On the other hand, decoder-only models, such as LLaMa~\cite{corr/abs-2302-13971}, ChatGPT~\cite{corr/abs-2302-04023}, and GPT-4~\cite{corr/abs-2303-12712}, are autoregressive models which are generative and are pretrained on the task of predicting the next word. The rest of the subsection focuses on how these two categories of models have been utilized in the context of KG completion.

One of the pioneers of LLM-based approaches for KG completion is KG Bidirectional Encoder Representations from Transformer (KG-BERT)~\cite{yao2019kg} which is fine-tuned on the task of KG completion and represents the triple as textual sequences. It takes the entity-relation-entity sequence and computes the scoring function using KG-BERT. The model represents entities and relations by their names or descriptions and takes the name/description word sequences as the input sentence of the BERT model for fine-tuning. 
Despite being an LLM-based approach, KG-BERT does not outperform models considering the structural information of a KG in terms of the ranking-based metrics, i.e., \textit{hits@k}. 
Kim et al.~\cite{KimHKS20} associates the shortcomings of KG-BERT with the fact that KG-BERT does not handle relations properly and it has difficulties in choosing the correct answer in the presence of lexically similar candidates. 
Therefore, Kim et al. propose a multitask learning based model~\cite{KimHKS20} that combines the task of relation prediction and relevance ranking with LP to lead to better learning of the relational information.

The previously described methods still suffer from high overheads because of costly scoring functions and a lack of structured knowledge of the textual encoders. A Structured Augmented text Representation (StAR) model~\cite{WangSLZW021} was proposed where each triple is partitioned into two asymmetric parts, similar to a translation-based graph embedding approach. Both parts were encoded into contextualized representations with the help of a Siamese-style textual encoder. 
Yet, KG completion methods based on pretrained language models (PLM) lag in performance w.r.t. structure-based algorithms. Lv et al.~\cite{LvL00LLLZ22} highlight that the reason underlying this lag is the evaluation setting, which is currently based on a Closed World Assumption (CWA), where an absent fact in the KG is considered false. 
Under the CWA, the performance of an LP algorithm is measured by its capability to predict a set of links that were removed from the KG. 
In contrast, LLMs introduce external knowledge which may lead to the prediction of new links that are semantically correct, but which were not in the original KG (and therefore not in the evaluation set) and, hence, not counted in the success metric. 
Additionally, the pretrained language models (PLMs) are utilized in an inappropriate manner, i.e., when triples are used as sentences it leads to incoherence in the generated sentences. 
Therefore, Lv et al.~\cite{LvL00LLLZ22} present a PLM-based approach dubbed PKGC. 
This work targets the first problem by proposing manual annotation as an alternative. However, a medium-sized dataset containing 10,000 entities and 10,000 triples in the test set will lead to true labels of at most 200 million triples precluding human annotation. This observation leads to a new evaluation metric called $CR@1$, where the triples are sampled from the test set and the missing entities are filled with the top-1 predicted entity. 
Manual annotation is then performed to measure the correct ratio of these triples. 
PKGC then addresses the second problem by converting each triple and its support information (i.e., the definition and the attribute information) into natural prompt sentences, which are fed to the PLM. 
PKGC outperforms structural and LLM-based methods in various modalities, i.e., with the attribute and definition.


GenKGC~\cite{XieZLDCXCC22} converts the KG completion task to a sequence-to-sequence (Seq2Seq) generation task, where the input to the model is $\langle head,~relation,~?\rangle$. Following the ``in-context learning" paradigm of GPT-3, the authors select some samples consisting of several triples with the same relation as in the input from the training set which is termed a relation-guided demonstration. It introduces an entity-aware hierarchical decoder during generation for better representation learning and reduced time complexity. On traditional datasets such as WN18RR and FB15k-237, GenKGC underperforms as compared to structural SotA models on \textit{hits@k} (confirming the hypothesis made By Lv et al.~\cite{LvL00LLLZ22}) and outperforms masked language model based approaches demonstrating the capabilities of generative models for KG completion. The Seq2Seq Generative Framework KG-Completion algorithm (KG-S2S)~\cite{ChenWLL22} takes into account the aspect of emerging KGs. Given a KG query, KG-S2S directly generates the target entity text by fine-tuning the pre-trained language model. KG-S2S learns the input representations of the entities and relations using entity descriptions, soft prompts, and Seq2Seq dropout. The KG elements, i.e., entity, relations, and timestamp are considered flat text, enabling the model to generate novel entities for KGs. The method is further evaluated in static, few-shot, and temporal settings.


\paragraph{Discussion.} 
LLMs have revolutionized the natural language processing (NLP) landscape by their ability to generate coherent and relevant text. However, they are not without limitations. One of the major concerns with LLMs is their tendency to hallucinate~\cite{SuchanekL23}, which means they may generate coherent yet entirely fictional information or associations. This poses a significant challenge in knowledge graph completion tasks, where the goal is to accurately predict missing links or facts. 

Moreover, LLMs typically require extensive fine-tuning and adaptation to specific domains to perform effectively. This is prevalent in specialized domains such as biomedical, legal, or scientific fields, where terms and context can vary significantly from general language use. 
The LLM performance is also lagging in torso and tail entities, where the LLM has limited or sparse knowledge about certain entities during learning~\cite{DBLP:journals/corr/abs-2308-10168}. 
Therefore, while LLMs show promise in improving the quality of KG completion, they need to be augmented with additional information to cope with knowledge scarcity or with domain-specific resources to provide accurate and reliable results in such domains.

Regarding the evaluation of KG completion methods, it is important to recognize that existing benchmarking methodologies often rely on the Closed World Assumption (CWA).
However, CWA-based evaluations may not be adequate for assessing methods that integrate external knowledge into the prediction process, such as LLMs.
This is because such approaches could produce accurate predictions that may not be represented in the KG itself, leading to spurious false positives under the CWA. 
Thus, there is a growing need for novel experimental configurations and evaluation metrics that can accommodate more complex and nuanced assessments, such as considering the confidence or uncertainty levels of predictions or incorporating additional external knowledge sources to provide context.

In conclusion, while LLMs present a valuable resource for knowledge graph completion tasks, their inherent limitations, particularly in terms of hallucination and domain-specific adaptation, need to be addressed. Additionally, the development of more sophisticated evaluation methodologies that go beyond the CWA will be crucial to ensure a reliable understanding of the LLM capabilities for KG completion in various domains.

\section{Towards Capturing Semantics in Knowledge Graph Embeddings}\label{sec:semantic-kg-embedding}

\subsection{Capturing Type Information for Knowledge Graph Completion} \label{sec:type-hierarchy}

Recent initiatives have been taken for leveraging the schematic information in the form of type information about entities, both with and without considering the type hierarchies. TKRL~\cite{XieLS16} considers the hierarchical information of the entity types by using hierarchical type encoders. It is based on the assumption that each entity should have multiple representations for its different (super)types. TransT~\cite{MaDJWG17} also proposes an approach that considers the entity type and its hierarchical structure. It goes one step further and constructs relation types from entity types and captures the prior distribution of the entity types by computing the type-based semantic similarity of the related entities and relations. 
Based on the prior distribution, multiple embedding representations of each entity (set of semantic vectors instead of a single vector) are generated in a different context and then the posterior probability of the entity and the relation prediction is estimated. Zhang et al.~\cite{ZhangKWM18} consider entity types as a constraint on the set of all the entities and let these type constraints induce an isomorphic collection of subsets in the embedding space. The framework introduces additional cost functions to model the fitness between these constraints and the entity and relation embeddings. JOIE~\cite{HaoCYSW19} employs cross-view and intra-view modeling, such that (i)~the cross-view association jointly learns the embeddings of the ontological concepts and the instance entities, and (ii)~the intra-view models learn the structured knowledge related to entities as well as the ontological information (the hierarchy-aware encoding) separately. The model is evaluated on the task of triple completion and entity typing. Another proposed method that learns both entity, relation, and entity type embeddings from entity-specific triples and type-specific triples is Automated Entity Type Representation for KG Embedding (AutoETER) for LP tasks~\cite{NiuLZPL20}. It considers the relation as a translation between the types of the head and the tail entity with the help of the relation-aware projection mechanism. 
Type-Aware Graph Attention neTworks for reasoning over KGs (TAGAT)~\cite{WangWHLG21} is one of the methods that combine the entity type information with the neighborhood information of the types while generating embeddings with the help of Graph ATtention networks (GAT). The relation level attention aims at distinguishing the importance of each associated relation for the entity. The neighborhood information of each type is also considered with the help of type-level attention since each relation may connect different entity groups even if the head entity belongs to the same group. Entity-level attention aims at determining how important each neighboring entity is for an entity under a given relation. TrustE~\cite{ZhaoLDXL21} is yet another method that aims at building entity-typed structured embeddings with tuple trustworthiness by taking into account possible noise in entity types which is usually ignored by the current methods. TrustE encodes the entities and entity types into separate spaces with a structural projection matrix. The trustworthiness is ensured by detecting the noisy entity types where the energy function focuses more on the pairs of an entity and its type with high trustworthiness. The model is evaluated by detecting entity-type noise as well as entity-type prediction.  


\begin{table}[t!]
\caption{Comparison of the expressivity of semantically enriched embeddings. Notation: $a,b$ are instances, $C,D,E$ are concepts, $r, r_1, r_2, s$ are relations. The symbol \cmark (resp. (\cmark) indicates that it has been demonstrated by construction or empirically that the approach supports (resp. partially) the expression; otherwise, \xmark is indicated. }
\label{tab:semantic_embeddings}
\footnotesize
\begin{tabular}{llccccccccc}
\hline
                                & Expressions                     & ELEm           & EMEL++         & BoxEL        & ELBE           & Box$^2$EL    & CatE           & OWL2Vec*    & TransOWL & OntoZSL \\ \hline
\multirow{5}{*}{Constructors}                 & $\bot$                          & \cmark   & \cmark   & \cmark & \cmark   & \xmark     & \xmark       & \xmark    & \xmark & \xmark\\
                                     & $\top$                          & \xmark       & \xmark       & \xmark     & \xmark       & \xmark     & \cmark   & \cmark & \xmark & \xmark\\
                                     & $\{a\}$                           & (\cmark) & \xmark       & \cmark & \xmark       & \cmark & \xmark       & \xmark    & \cmark & \xmark\\
                                     & $C(a)$                          & \cmark   & \cmark   & \cmark & \cmark   & \cmark & \xmark       & \cmark  & \cmark & \cmark \\
                                     & $r(a,b)$                        & \cmark   & \cmark   & \cmark & \cmark   & \cmark & \xmark       & \cmark  & \cmark & \cmark \\ \hline
\multirow{7}{*}{$\mathcal{EL}^{++}$} & $C \sqsubseteq D$               & \cmark   & \cmark   & \cmark & \cmark   & \cmark & \cmark   & \cmark  & \cmark & \cmark \\
                                     & $C \sqcap D \sqsubseteq E$      & (\cmark) & \cmark   & \cmark & \cmark   & \cmark & (\cmark) & \xmark    & \xmark & \xmark\\
                                     & $\exists r.C \sqsubseteq D$     & \cmark   & \cmark   & \cmark & \cmark   & \cmark & \cmark   & \cmark & \cmark & \cmark\\
                                     & $C  \sqsubseteq \exists r.D$    & \cmark   & \cmark   & \cmark & \cmark   & \cmark & \cmark   & \cmark & \cmark & \xmark\\
                                     & $C \sqcap D \sqsubseteq \bot$   & \xmark       & \cmark   & \cmark & \cmark   & \cmark & \cmark   & \xmark   &  \xmark  & \xmark\\
                                     & $\exists r.C \sqsubseteq \bot $ & (\cmark) & (\cmark) & \cmark & (\cmark) & \cmark & \xmark       & \xmark   &  \xmark  & \xmark\\
                                     & $ C  \sqsubseteq \bot $         & \xmark       & \cmark   & \cmark & \cmark   & \cmark & \xmark       & \xmark    & \xmark & \xmark \\ \hline
\multirow{5}{*}{$\mathcal{ALC}$}     & $C \sqcup D$                    & \xmark       & \xmark       & \xmark     & \xmark       & \xmark     & \cmark   & \xmark     & \xmark & \xmark \\
                                     & $\forall r.D$                   & \xmark       & \xmark       & \xmark     & \xmark       & \xmark     & \cmark   & \xmark   &  \xmark & \xmark \\
                                     & $\neg C$                        & \xmark       & \xmark       & \xmark     & \xmark       & \xmark     & \cmark   & \xmark   &  \xmark & \xmark \\
                                     & $C \sqsubseteq \forall r.D$     & \xmark       & \xmark       & \xmark     & \xmark       & \xmark     & \xmark       & \cmark & \cmark & \cmark \\
                                     & $\forall r.C \sqsubseteq D$     & \xmark       & \xmark       & \xmark     & \xmark       & \xmark     & \xmark       & \cmark & \cmark & \xmark \\
\midrule
\multirow{2}{*}{$\mathcal{SH}$}      & $r \sqsubseteq s$               & \xmark       & \cmark   & \xmark     & \xmark       & \xmark     & \xmark       & \cmark & \cmark & \cmark \\
                                     & $r_1 \circ r_2 \sqsubseteq s$   & \xmark       & \cmark   & \xmark     & \xmark       & \xmark     & \xmark       & \cmark & \xmark & \xmark \\ 
\bottomrule
\end{tabular}
\end{table}
\subsection{Ontology-Enriched Knowledge Graph Embeddings}
\label{sec:semantic-embeddings}

Even though the previously discussed approaches use some of the schematic information, such as the types of entities and the type hierarchy, they still ignore much of 
the concept level knowledge captured in ontologies, i.e., TBox information in terms of description logic axioms. 
As reported by Chen et al.~\cite{ChenHJHAH21}, KGE that interpret TBox triples as ABox triples (i.e., ignoring their semantics) do not achieve high performance in reasoning tasks such as membership and subsumption prediction.
Therefore, in this section, we analyse the KGEs that do take TBox information into account, and their expressivity in regards to the constructs and axioms of logic languages discussed in Section~\ref{sec:preliminaries}. 
Table~\ref{tab:semantic_embeddings} presents an overview of these approaches.

A first generation of these systems represented concepts as (high-dimensional) spheres in the embedding space (e.g. \cite{KulmanovLYH19}). However, while the intersection of concepts is a common operation, the intersection of two n-balls is not an n-ball, leading to challenges when measuring the distance between concepts and inferring equivalence between concepts. A second generation instead represents concepts as high-dimensional boxes, since boxes are closed under intersection. 
ELEm~\cite{KulmanovLYH19} is one of the first of these approaches and generates low dimensional vector spaces from $\mathcal{EL}^{++}$ by approximating the interpretation function by extending TransE with the semantics of conjunction, existential quantification, and the bottom concept.  
It is evaluated for LP based on protein-protein interaction. 
EMEL++~\cite{MondalBM21} evaluates the algorithm on the task of subsumption reasoning and compares it to the ELEm where these semantics are represented but not properly evaluated.
Similar to TransE, EMEL++ interprets relations as the translations operating between the classes. 
BoxEL~\cite{XiongPTNS22} and ELBE~\cite{abs-2202-14018} extend ELEm by representing concepts as axis parallel boxes with two vectors for either the lower and upper corners or the center and offset. In BoxEL~\cite{XiongPTNS22}, the authors show the aforementioned advantage of box embeddings over ball embeddings with the help of an example related to conjunction operator, i.e., the ball embeddings cannot express $Parent \sqcap Male \equiv Father$ as properly as compared to the box embeddings. Moreover, the translations cannot model the relation $isChildOf$ between a $Person$ and a $Parent$ when they have two different volumes. In addition to the box representation, ELBE~\cite{abs-2202-14018} defines several loss functions for each of the normal forms representing the axioms expressed in $\mathcal{EL}^{++}$ (shown in Table~\ref{tab:semantic_embeddings}) such as conjunction, the bottom concept, etc. Taking a further step, Box$^2$EL~\cite{abs-2301-11118} learns representations of not only the concepts but also the roles as boxes for preserving as much semantics of the ontology as possible. 
It uses a similar mechanism to BoxEL for the representation of the concepts. The previous methods define roles (binary relations) as translations as in TransE but Box$^2$EL associates every role with a head box and a tail box so that every point in the head box is related to every point in the tail box with the help of bump vectors. Bump vectors model interaction between the concepts and dynamically move the position of the embeddings of related concepts.
CatE~\cite{abs-2305-07163} on the other hand embeds $\mathcal{ALC}$ ontologies with the help of category theoretical semantics, i.e., the semantics of logic languages that formalizes interpretations using categories instead of sets. This is advantageous because categories have a graph-like structure. 
TransOWL~\cite{TransOWL} and its extensions allow for the inclusion of OWL axioms into the embedding process by modifying the loss function of TransE so that it gives a higher weight to triples that involve OWL vocabulary such as \texttt{owl:inverseOf}, \texttt{owl:equivalentClass}, etc. OWL2Vec*~\cite{ChenHJHAH21} uses a word embedding model to create embeddings from entities and words from the generated corpus. This corpus is generated from random walks over the ontology. The method is evaluated on the task of class membership prediction and class subsumption prediction. OntoZSL~\cite{GengC0PYYJC21} is another approach that takes into account the ontological schema by considering the structural information as well as the textual information of the classes (via \texttt{rdfs:subClassOf}) and predicates (via \texttt{rdfs:subPropertyOf}, \texttt{rdfs:domain}, and \texttt{rdfs:range}) in the ontology for KG completion.  

\paragraph{Discussion.}
This section discussed the methods leveraging schematic information contained in the KGs starting from type hierarchy to the semantics of classes and predicates in the description logic axioms. 
The approaches utilizing type hierarchies for LP are, however, in most cases evaluated on the traditional triple and entity-type prediction tasks but completely ignore tasks related to schema completion involving terms only from the TBox in the KG. Furthermore, the limited expressivity of these models makes them unsuitable for the more complex tasks of deductive reasoning. These shortcomings are addressed by the approaches representing the semantics underlying different description logic axioms with the help of box embeddings. Yet, these approaches are limited to certain kinds of axioms and do not cover more expressive statements like role transitivity. 
This is evidenced in Table~\ref{tab:semantic_embeddings} where only a few approaches can handle certain aspects of $\mathcal{SH}$, yet the expressivity of OWL ontologies (from $\mathcal{SHIF^{(D)}}$ to $\mathcal{SROIQ^{(D)}}$) is far to be reached.
Lastly, the approaches discussed in this section still lack unified evaluation w.r.t. benchmark datasets leading to unclear performance comparison of the existing algorithms and of the impact of adding such schematic information for training the models.

\section{Existing Evaluation Settings for Knowledge Graph Completion using Embeddings} \label{sec:evaluation}

This section provides an overview of the evaluation protocol including the benchmark datasets as well as the used evaluation metrics. Table~\ref{table:evaluation_protocol} shows an overview of the evaluation tasks based on the categories of algorithms introduced in the rest of the paper along with the benchmark datasets.

\paragraph{Benchmark Datasets.}

The benchmark datasets vary from task to task, depending on the requirements of the algorithms. Most datasets are based on WordNet, YAGO, and Freebase which utilize only the triple information and do not incorporate the ontology by default. 
A detailed comparison of the datasets for LP in a transductive setting is provided by Gesese et al.~\cite{gesese2021LitWD}. 
The datasets used for link prediction (LP) typically provide training/validation sets; still, these splits are computed arbitrarily. Therefore, a fine-grained analysis of how the methods perform on different relations or classes is not always possible with these datasets.
ELEm has been evaluated on the Protein-Protein Interaction (PPI) dataset for the LP task. However, the successor algorithms introduced datasets that consider more expressive description logic axioms such as Gene Ontology (GO)~\cite{KulmanovH17} or  FoodOn~\cite{Dooley2018FoodOnAH}.
In addition, some works~\cite{MondalBM21,DBLP:conf/semweb/XiongPTNS22} for ontology embedding models have further classified the test sets into normal forms (NF) to evaluate the models on subsumption prediction between atomic concepts (NF1), atomic concepts and conjunctions (NF2), and atomic concepts and existential restrictions (NF3 and NF4).

\begin{table}[t] 
\centering
\caption{Overview of the tasks and the datasets used for evaluation for each of the categories.}
\begin{tabular}{p{3.5cm}|p{5.5cm}|p{5.25cm} } 
\toprule
\textbf{Category}& \textbf{Tasks}&  \textbf{Datasets}\\ \midrule

\multirow{2}{*}{Transductive Link Prediction} & Link Prediction  & FB15K~\cite{bordes2011learning}, FB15K-237~\cite{toutanova-chen-2015-observed}, WN18~\cite{bordes2011learning}, WN18RR~\cite{dettmers2018convolutional}, YAGO3-10~\cite{dettmers2018convolutional}, LiterallyWikidata~\cite{GeseseAS21} \\ \cline{2-3}

& Entity Classification, Triple Classification  & FB15K~\cite{bordes2011learning}, FB15K-237~\cite{toutanova-chen-2015-observed}, WN18~\cite{bordes2011learning}, WN18RR~\cite{dettmers2018convolutional}, YAGO3-10~\cite{dettmers2018convolutional}\\ \hline

Inductive Link Prediction &Link Prediction &Wikidata5M~\cite{Wang2021KEPLERAU}, Wikidata68K~\cite{GeseseSA22}, ILPC 2022~\cite{abs-2203-01520} \\ \hline

LLM-Based Methods & Link Prediction & FB15K237, WN18RR, FB15K-237N~\cite{LvL00LLLZ22}, FB15K-237NH~\cite{LvL00LLLZ22}, Wiki27K~\cite{LvL00LLLZ22}, OpenBG500~\cite{XiongYCGW18}, Nell-One~\cite{XiongYCGW18}  \\ \hline
\multirow{2}{*}{Methods using Type-Hierarchy}& Link Prediction& WN18RR, WN18, FB15k, FB15KET, YAGO43KET~\cite{10.1145/3132847.3133095}\\ \cline{2-3}
 & Entity Type Prediction& WN18RR, WN18, FB15k, FB15KET, YAGO43KET~\cite{10.1145/3132847.3133095}, FB15K+~\cite{xie2016representation}, FB15K*, JF17K~\cite{ZhangKWM18}, NELL-995~\cite{WangWHLG21},  YAGO26K-906~\cite{Xie2016DKRL}, DB111K-174~\cite{Xie2016DKRL} \\ \hline

\multirow{2}{*}{\parbox{3cm}{Methods using Description Logic Axioms}}& Inductive Reasoning (Link Prediction, Subsumption, Class Membership) & 
NELL-ZS~\cite{qin2020generative}, Wikidata-ZS~\cite{qin2020generative}, GO~\cite{KulmanovH17},  FoodOn~\cite{Dooley2018FoodOnAH}  \\ 

\cline{2-3}

 & Deductive Reasoning (Unsatisfiability of the named concept, predicting axioms in deductive closure of an ontology)  & 
  FoodOn~\cite{Dooley2018FoodOnAH}, HeLIS~\cite{DragoniBME18}, GO~\cite{KulmanovH17}, SNOMED-CT~\cite{Donnelly2006SNOMEDCTTA}, GALEN~\cite{Pole1996TheGH}, Anatomy~\cite{Mungall2012UberonAI}, PPI~\cite{SzklarczykMC0WS17}, FOBI~\cite{abs-2305-07163}, NRO~\cite{abs-2305-07163}\\



\bottomrule
\end{tabular}
\label{table:evaluation_protocol}
\end{table}

\paragraph{Metrics.}
For the task of LP, evaluation protocols typically involve rank-based metrics. Most of the algorithms use \textit{hits@k} where $k$ varies from 1 to 100, and typically values are 1, 3, 10, 50. Other ranking metrics are Mean Rank (MR) and Mean Reciprocal Rank (MRR).
Besides rank-based metrics, for the task of entity classification or entity type prediction, some works (e.g., \cite{weller2021predicting}) report on metrics used for information retrieval, i.e., precision, recall, and their corresponding aggregates such as F1 scores. 
In addition, works on ontology embedding models~\cite{abs-2305-07163,KulmanovLYH19,DBLP:conf/semweb/XiongPTNS22} also report on the area under the curve 
(AUC) of the ROC curve.

\paragraph{Evaluation and Reported Results.}

Most of the algorithms are evaluated on the task of LP since they consider only the triple structure in the KG. Exceptions to this are the tasks that are used to evaluate the algorithms, incorporating description logic axioms. 
In the case of transductive LP, studies have shown that the efficiency of these algorithms greatly depends on the choice of hyperparameters. 
For example, Ruffinelli et al.~\cite{RuffinelliBG20} show that RESCAL, one of the earliest models, outperforms all the later approaches by training the models with proper hyperparameters. A similar study by Ali et al.~\cite{AliBHVGSFTL22} performs a large-scale evaluation of 21 KG embedding models using PyKeen\footnote{\url{https://pykeen.readthedocs.io/en/stable/}} 
and provides the best set of hyper-parameters. The results of the study show that the performance of the models is not only determined by their architecture. The training approach, loss function, and the explicit modeling of inverse relations are also crucial. This leads to the conclusion that a unified benchmarking is needed which allows for comparing the approaches under the same conditions. This will reveal interesting insights, e.g., some models considered not competitive anymore can still outperform more sophisticated approaches under certain configurations. While the evaluation setting for transductive LP algorithms has been well established, the other three categories still lack a unified evaluation setting.
Lastly, most of the evaluations present reports on the models' performance over the entire test set. These analyses fail to provide a detailed understanding of the limitations of the embeddings on different aspects of the KG, e.g., tail elements (entities, classes, or relations) or relation types (e.g., transitive, symmetric, etc.). Few studies have addressed this issue and present specific results, e.g., per normal form in the ontology~\cite{abs-2301-11118}, or aggregate metrics like F1-macro~\cite{weller2021predicting} that take into account the weights of classes. Still, there is a notable discrepancy in the literature regarding the manner datasets, benchmarks, and analyses are reported. 

\section{Conclusion and Recommendations} \label{sec:conclusion}

In this section, we take a critical perspective on the SoTA over the past decade, based on the overview given in the preceding sections. We end with recommendations for fruitful future directions of work. 

\subsection{Critical Reflection}

\paragraph{Ignoring semantics.}
The majority of KG completion algorithms only consider the ABox (Definition \ref{def:triple-set}). This effectively reduces a KG to a ``data graph", i.e., relations between entities. 
Only recent algorithms have appeared that treat KGs as objects with computer-interpretable semantics (Definition \ref{def:semantic-kg}), such as those discussed in Section \ref{sec:type-hierarchy} (to a limited extent) and Section \ref{sec:semantic-embeddings} (to a larger extent). 

\paragraph{Limitations of the transductive setting.}
The majority of the work on KG completion follows a transductive setting, i.e., not allowing for the completion of a KG with new entities or relations. Only a few algorithms set in an inductive setting consider only the prediction of new entities, with almost no algorithms aiming to predict new relations. Limiting algorithms to the transductive setting severely restricts the downstream tasks to which these algorithms can be applied. For example, missing relations between known proteins and drugs can be predicted (protein-drug interaction), but transductive algorithms cannot be used to predict new drugs from the knowledge of known proteins. 

\paragraph{Evaluation settings.}
The community working on KG completion is severely hampered by a lack of standardised evaluation settings. There are multiple concerns in this area: 
\begin{itemize}     
\item \textit{No standard protocols for hyper-parameter sweeping}, leading to incomparable and unreproducible results. This led to the rather embarrassing result that a paper from 2020 \cite{RuffinelliBG20} threw into doubt the progress of the entire field over a decade since the early work on RESCAL in 2011 \cite{nickel2011three}. 

\item \textit{Evaluation datasets with serious limitations}. It took years before the widely used FB15K dataset was discovered to suffer from serious data leakage from the training to testing and validation splits because many triples were simply inverses of other triples, allowing trivial inverse-triple prediction algorithms to already gain a high score~\cite{toutanova-chen-2015-observed}. Furthermore, the small number of datasets used in the literature (Table \ref{table:evaluation_protocol}) carries the risk that methods will optimise for the benchmark, instead of optimise for the task. 

\item \textit{Dataset size.}
Datasets used in evaluations are typically small (e.g. on the order of $10^5$ triples for the widely used FB15k-237 and YAGO3-10) in comparison to the KGs that are currently in routine practical use (often on the order of $10^8$ triples or beyond). 

\item \textit{Evaluation metric.}
The most widely deployed evaluation metric is \emph{hits@k} on a held-out set of triples. This method favours reconstructing known links (that were already present in the original KG, but held out) over the prediction of genuinely new links that are semantically correct but did not appear in the original KG. A method may produce many genuinely new links that rank higher than a predicted held-out link, and such a high-value method would end up with a low \emph{hits@k} score. Ironically, it is these methods which are likely to be of most value in downstream tasks. The recently proposed metric \emph{sem@k} \cite{DBLP:journals/corr/abs-2301-05601} aims to rectify this to some extent, although it remains unclear at the moment how much of this evaluation can be done without expensive human annotation of gold standards. 
\end{itemize}

\paragraph{Bias against external knowledge.} Almost all current KG completion methods aim to predict links in a KG based on the properties of the KG itself, and this even holds true for most of the inductive methods that go beyond a transductive setting. 
On the other hand, the new generation of methods that exploit LLMs for KG prediction uses an external knowledge source (i.e., LLMs, see Section~\ref{sec:llm-kg-completion}) to predict new links. Not only is this an obviously promising step to take in an inductive setting, but it also holds the promise of taking LP beyond simply extrapolating the structural patterns that are already present in the KG. After all, methods that merely extrapolate the existing patterns in a KG are likely to simply replicate the sources of the incompleteness of the KG and are thus unlikely to actually solve the problem that was motivating the KG completion task in the first place. It is important to observe that LLMs are by no means the only useful source of external knowledge that can be injected into the KG completion process. Other KGs can also provide such background knowledge, leading to an interesting blurring between the tasks of KG linking and KG completion\textcolor{blue}{~\cite{BaumgartnerDPB23}}. Recent work on exploiting the temporal evaluation of a KG as the source of information is another example of using information outside the KG for KG completion. 

\subsection{Recommendations}

The above critical reflections lead to several recommendations for future directions. 

\paragraph{Towards Semantic Embeddings.} If the work on \emph{knowledge graph} completion is to go beyond simply \emph{data graph} completion, the effort will need to focus on including the semantics of the KG. This can be as simply accounting for known relations between types such as the subsumption hierarchy or known disjointness relations between types to more sophisticated reasoning about the algebraic properties of roles (symmetry, anti-symmetry, transitivity, etc.), or properties such as minimal and maximal cardinality of roles. The methods discussed in Section~\ref{sec:semantic-embeddings} still need to be extended to take into account these semantics.

\paragraph{Use of External Resources.} 
We also believe that this will have to go hand in hand with a willingness to develop KG completion algorithms that take into account knowledge sources beyond the original KG. These knowledge sources can be (i) the updated versions of the KG themselves leading to the necessity to build dynamic KG embedding algorithms, (ii) building KG embedding algorithms containing temporal information about facts making it more precise and updatable for considering time-sensitive facts, (iii) rules to encode domain-specific constraints and capture complex relationships and structural patterns in the KG that may not be expressed with logical axioms as the ones presented in Table~\ref{table:evaluation_protocol},  (iv)  using multiple knowledge graphs to integrate more complete knowledge about the facts represented in a KG, and (v) last but not least, LLMS which contain a vast amount of background knowledge. 

\paragraph{Datasets and Evaluation.} Evaluation datasets need to grow larger in size and more diverse in nature, and perhaps be more representative of real-world downstream tasks. This goes hand-in-hand with the first two recommendations. More methodological hygiene needs to be practiced in both the protocols for evaluations (e.g. hyper-parameter settings), as well as the reporting about these settings. These future recommendations come with different kinds of challenges such as larger datasets on which the embeddings should be computed needing huge computational resources along with hallucination problems as is the case with LLMs. In addition to the dataset and experimental settings, alternative metrics will need to be developed that better reflect the actual value of the predicted links for downstream tasks.

\bibliographystyle{ios1}           
\bibliography{bib-new}        

%

\end{document}